\documentclass[12pt,final]{elsarticle}
\usepackage[latin9]{inputenc}
\usepackage{array}
\usepackage{verbatim}
\usepackage{amsmath}
\usepackage{amssymb}

\makeatletter

\providecommand{\tabularnewline}{\\}

\usepackage{enumitem}		

\usepackage{lineno}

\journal{Neurocomputing}
\usepackage{url}

\DeclareMathOperator*{\T}{\scriptscriptstyle \top}

\DeclareMathOperator*{\argmin}{arg\,min}

\def\smallunderbrace#1{\mathop{\vtop{\m@th\ialign{##\crcr
   $\hfil\displaystyle{#1}\hfil$\crcr
   \noalign{\kern3\p@\nointerlineskip}%
   \small \upbracefill\crcr\noalign{\kern3\p@}}}}\limits}

\newcommand\undermat[2]{%
  \makebox[0pt][l]{$\smash{\smallunderbrace{\phantom{%
    \begin{matrix}#2\end{matrix}}}_{\text{$#1$}}}$}#2}

\usepackage{tikz}
\renewcommand\thempfootnote{\fnsymbol{mpfootnote}}

\makeatother

\begin{document}
\begin{frontmatter}



\title{\texttt{PSDVec}: a Toolbox for Incremental and Scalable Word Embedding}


\author[label1]{Shaohua Li\corref{cor1}}

\ead{shaohua@gmail.com}

\author[label2]{Jun Zhu}

\ead{dcszj@tsinghua.edu.cn}

\author[label1]{Chunyan Miao}

\ead{ascymiao@ntu.edu.sg}

\address[label1]{Joint NTU-UBC Research Centre of Excellence in Active Living for
the Elderly (LILY), Nanyang Technological University, Singapore}

\address[label2]{Tsinghua University, P.R.China}

\cortext[cor1]{Corresponding author.}

\begin{abstract}
PSDVec is a Python/Perl toolbox that learns word embeddings, i.e.
the mapping of words in a natural language to continuous vectors which
encode the semantic/syntactic regularities between the words. PSDVec
implements a word embedding learning method based on a weighted low-rank
positive semidefinite approximation. To scale up the learning process,
we implement a blockwise online learning algorithm to learn the embeddings
incrementally. This strategy greatly reduces the learning time of
word embeddings on a large vocabulary, and can learn the embeddings
of new words without re-learning the whole vocabulary. On 9 word similarity/analogy
benchmark sets and 2 Natural Language Processing (NLP) tasks, PSDVec
produces embeddings that has the best average performance among popular
word embedding tools. PSDVec provides a new option for NLP practitioners.
\end{abstract}
\end{frontmatter}



\section{Introduction}

Word embedding has gained popularity as an important unsupervised
Natural Language Processing (NLP) technique in recent years. The task
of word embedding is to derive a set of vectors in a Euclidean space
corresponding to words which best fit certain statistics derived from
a corpus. These vectors, commonly referred to as the \textit{embeddings},
capture the semantic/syntactic regularities between the words. Word
embeddings can supersede the traditional one-hot encoding of words
as the input of an NLP learning system, and can often significantly
improve the performance of the system.

There are two lines of word embedding methods. The first line is neural
word embedding models, which use softmax regression to fit bigram
probabilities and are optimized with Stochastic Gradient Descent (SGD).
One of the best known tools is word2vec\footnote{https://code.google.com/p/word2vec/}
\cite{word2vec}. The second line is low-rank matrix factorization
(MF)-based methods, which aim to reconstruct certain bigram statistics
matrix extracted from a corpus, by the product of two low rank factor
matrices. Representative methods/toolboxes include Hyperwords\footnote{https://bitbucket.org/omerlevy/hyperwords}
\cite{ppmi,tacl}, GloVe\footnote{http://nlp.stanford.edu/projects/glove/}
\cite{glove}, Singular\footnote{https://github.com/karlstratos/singular}
\cite{cca2}, and Sparse\footnote{https://github.com/mfaruqui/sparse-coding}
\cite{sparsecoding1}. All these methods use two different sets of
embeddings for words and their context words, respectively. SVD based
optimization procedures are used to yield two singular matrices. Only
the left singular matrix is used as the embeddings of words. However,
SVD operates on $\boldsymbol{G}^{\T}\boldsymbol{G}$, which incurs
information loss in $\boldsymbol{G}$, and may not correctly capture
the \emph{signed correlations }between words. An empirical comparison
of popular methods is presented in \cite{tacl}.

The toolbox presented in this paper is an implementation of our previous
work \cite{psdvec}. It is a new MF-based method, but is based on
eigendecomposition instead. This toolbox is based on \cite{psdvec},
where we estabilish a Bayesian generative model of word embedding,
derive a weighted low-rank positive semidefinite approximation problem
to the Pointwise Mutual Information (PMI) matrix, and finally solve
it using eigendecomposition. Eigendecomposition avoids the information
loss in based methods, and the yielded embeddings are of higher quality
than SVD-based methods. However eigendecomposition is known to be
difficult to scale up. To make our method scalable to large vocabularies,
we exploit the sparsity pattern of the weight matrix and implement
a divide-and-conquer approximate solver to find the embeddings incrementally. 

Our toolbox is named \emph{Positive-Semidefinite Vectors (PSDVec)}.
It offers the following advantages over other word embedding tools:
\begin{enumerate}[itemsep=-0.3ex]
\item The incremental solver in PSDVec has a time complexity $O(cd^{2}n)$
and space complexity $O(cd)$, where $n$ is the total number of words
in a vocabulary, $d$ is the specified dimensionality of embeddings,
and $c\ll n$ is the number of specified core words. Note the space
complexity does not increase with the vocabulary size $n$. In contrast,
other MF-based solvers, including the core embedding generation of
PSDVec, are of $O(n^{3})$ time complexity and $O(n^{2})$ space complexity.
Hence asymptotically, PSDVec takes about $cd^{2}/n^{2}$ of the time
and $cd/n^{2}$ of the space of other MF-based solvers\footnote{Word2vec adopts an efficient SGD sampling algorithm, whose time complexity
is only $O(kL)$, and space complexity $O(n)$, where $L$ is the
number of word occurrences in the input corpus, and $k$ is the number
of negative sampling words, typically in the range $5\sim20$.};
\item Given the embeddings of an original vocabulary, PSDVec is able to
learn the embeddings of new words incrementally. To our best knowledge,
none of other word embedding tools provide this functionality; instead,
new words have to be learned together with old words in batch mode.
A common situation is that we have a huge general corpus such as English
Wikipedia, and also have a small domain-specific corpus, such as the
NIPS dataset. In the general corpus, specific terms may appear rarely.
It would be desirable to train the embeddings of a general vocabulary
on the general corpus, and then incrementally learn words that are
unique in the domain-specific corpus. Then this feature of incremental
learning could come into play;
\item On word similarity/analogy benchmark sets and common Natural Language
Processing (NLP) tasks, PSDVec produces embeddings that has the best
average performance among popular word embedding tools;
\item PSDVec is established as a Bayesian generative model \cite{psdvec}.
The probabilistic modeling endows PSDVec clear probabilistic interpretation,
and the modular structure of the generative model is easy to customize
and extend in a principled manner. For example, global factors like
topics can be naturally incorporated, resulting in a hybrid model
\cite{topicvec} of word embedding and Latent Dirichlet Allocation
\cite{lda}. For such extensions, PSDVec would serve as a good prototype.
While in other methods, the regression objectives are usually heuristic,
and other factors are difficult to be incorporated.
\end{enumerate}

\section{Problem and Solution}

PSDVec implements a low-rank MF-based word embedding method. This
method aims to fit the $\textrm{PMI}(s_{i},s_{j})=\log\frac{P(s_{i},s_{j})}{P(s_{i})P(s_{j})}$
using $\boldsymbol{v}_{s_{j}}^{\T}\boldsymbol{v}_{s_{i}}$, where
$P(s_{i})$ and $P(s_{i},s_{j})$ are the empirical unigram and bigram
probabilities, respectively, and $\boldsymbol{v}_{s_{i}}$ is the
embedding of $s_{i}$. The regression residuals $\textrm{PMI}(s_{i},s_{j})-\boldsymbol{v}_{s_{j}}^{\T}\boldsymbol{v}_{s_{i}}$
are penalized by a monotonic transformation $f(\cdot)$ of $P(s_{i},s_{j})$,
which implies that, for more frequent (therefore more important) bigram
$s_{i},s_{j}$, we expect it is better fitted. The optimization objective
in the matrix form is
\begin{equation}
\boldsymbol{V}^{*}=\argmin_{\boldsymbol{V}}||\boldsymbol{G}-\boldsymbol{V}^{\T}\boldsymbol{V}||_{f(\boldsymbol{H})}+\sum_{i=1}^{W}\mu_{i}\Vert\boldsymbol{v}_{s_{i}}\Vert_{2}^{2},\label{eq:obj}
\end{equation}
where $\boldsymbol{G}$ is the PMI matrix, $\boldsymbol{V}$ is the
embedding matrix, $\boldsymbol{H}$ is the bigram probabilities matrix,
$||\cdot||_{f(\boldsymbol{H})}$ is the $f(\boldsymbol{H})$-weighted
Frobenius-norm, and $\mu_{i}$ are the Tikhonov regularization coefficients.
The purpose of the Tikhonov regularization is to penalize overlong
embeddings. The overlength of embeddings is a sign of overfitting
the corpus. Our experiments showed that, with such regularization,
the yielded embeddings perform better on all tasks.

\eqref{eq:obj} is to find a weighted low-rank positive semidefinite
approximation to $\boldsymbol{G}$. Prior to computing $\boldsymbol{G}$,
the bigram probabilities $\{P(s_{i},s_{j})\}$ are smoothed using
Jelinek-Mercer Smoothing.

A Block Coordinate Descent (BCD) algorithm \cite{weightedlow} is
used to approach \eqref{eq:obj}, which requires eigendecomposition
of $\boldsymbol{G}$. The eigendecomposition requires $O(n^{3})$
time and $O(n^{2})$ space, which is difficult to scale up. As a remedy,
we implement an approximate solution that learns embeddings incrementally.
The incremental learning proceeds as follows:
\begin{enumerate}[topsep=2pt,itemsep=-0.2ex]
\item Partition the vocabulary $\boldsymbol{S}$ into $K$ consecutive
groups $\boldsymbol{S}_{1},\cdots,\boldsymbol{S}_{k}$. Take $K=3$
as an example. $\boldsymbol{S}_{1}$ consists of the most frequent
words, referred to as the \textbf{core words}, and the remaining words
are \textbf{noncore words};
\item Accordingly partition $\boldsymbol{G}$ into $K\times K$ blocks as
$\left(\begin{array}{c|cc}
\boldsymbol{G}_{11} & \boldsymbol{G}_{12} & \boldsymbol{G}_{13}\\
\hline \boldsymbol{G}_{21} & \boldsymbol{G}_{22} & \boldsymbol{G}_{23}\\
\boldsymbol{G}_{31} & \boldsymbol{G}_{32} & \boldsymbol{G}_{33}
\end{array}\right).$ Partition $f(\boldsymbol{H})$ in the same way. $\boldsymbol{G}_{11},f(\boldsymbol{H})_{11}$
correspond to \textbf{core-core bigrams} (consisting of two core words).
Partition $\boldsymbol{V}$ into $\begin{array}{c}
\begin{pmatrix}\undermat{\boldsymbol{S}_{1}}{\boldsymbol{V}_{1}} & \negthickspace\undermat{\boldsymbol{S}_{2}}{\;\boldsymbol{V}_{2}} & \negthickspace\undermat{\boldsymbol{S}_{3}}{\;\boldsymbol{V}_{3}}\end{pmatrix}\\
\rule{0pt}{15pt}
\end{array}$;
\item For core words, set $\mu_{i}=0$, and solve $\argmin_{\boldsymbol{V}}||\boldsymbol{G}_{11}-\boldsymbol{V}_{1}^{\T}\boldsymbol{V}_{1}||_{f(\boldsymbol{H}_{1})}$
using eigendecomposition, obtaining core embeddings $\boldsymbol{V}_{1}^{*}$;
\item Set $\boldsymbol{V}_{1}=\boldsymbol{V}_{1}^{*}$, and find $\boldsymbol{V}_{2}^{*}$
that minimizes the total penalty of the $12$-th and 21-th blocks
(the 22-th block is ignored due to its high sparsity):
\[
\argmin_{\boldsymbol{V}_{2}}\Vert\boldsymbol{G}_{12}-\boldsymbol{V}_{1}^{\T}\boldsymbol{V}_{2}\Vert_{f(\boldsymbol{H})_{12}}^{2}+\Vert\boldsymbol{G}_{21}-\boldsymbol{V}_{2}^{\T}\boldsymbol{V}_{1}\Vert_{f(\boldsymbol{H})_{21}}^{2}+\sum_{s_{i}\in\boldsymbol{S}_{2}}\mu_{i}\Vert\boldsymbol{v}_{s_{i}}\Vert^{2}.
\]
The columns in $\boldsymbol{V}_{2}$ are independent, thus for each
$\boldsymbol{v}_{s_{i}}$, it is a separate weighted ridge regression
problem, which has a closed-form solution \cite{psdvec};
\item For any other set of noncore words $\boldsymbol{S}_{k}$, find $\boldsymbol{V}_{k}^{*}$
that minimizes the total penalty of the $1k$-th and $k1$-th blocks,
ignoring all other $kj$-th and $jk$-th blocks; 
\item Combine all subsets of embeddings to form $\boldsymbol{V}^{*}$. Here
$\boldsymbol{V}^{*}=(\boldsymbol{V}_{1}^{*},\boldsymbol{V}_{2}^{*},\boldsymbol{V}_{3}^{*})$.
\end{enumerate}

\section{Software Architecture and Functionalities }

Our toolbox consists of 4 Python/Perl scripts: \texttt{extractwiki.py},
\texttt{gramcount.pl}, \texttt{factorize.py }and \texttt{evaluate.py}.
Figure 1 presents the overall architecture.

\begin{figure}
\noindent \centering{}\usetikzlibrary{shapes.geometric, arrows} \tikzstyle{io} = [trapezium, trapezium left angle=75, trapezium right angle=105, minimum width=2cm, minimum height=1cm, text centered, draw=black, outer sep=1pt, inner xsep=-4pt, fill=blue!30] \tikzstyle{process} = [rectangle, minimum width=3cm, minimum height=1cm, text centered, draw=black, outer sep=0, inner sep=5pt, fill=orange!30, rounded corners ] \tikzstyle{decision} = [diamond, minimum width=1cm, minimum height=1cm, text centered, inner sep=-2pt, draw=black, fill=green!30] \tikzstyle{arrow} = [thick,->,>=stealth] \begin{tikzpicture}[node distance=2cm, scale=0.6, transform shape] \node (corpus) [io] {Corpus}; \node (gramcount) [process, right of=corpus, xshift=2cm, align=center] {extractwiki.py \\ gramcount.pl}; \node (bigram) [io, below of=gramcount, yshift=0.3cm] {Bigram Statistics}; \node (factorize) [process, below of=bigram, yshift=0.3cm, align=center] {factorize.py: we\_factorize\_EM() \\ Factorize core block}; \node (core) [io, right of=factorize, xshift=4cm] {Core Embeddings}; \node (more) [decision, below of=factorize, yshift=-0.7cm,text width=2cm] {More noncore words?}; \node (concat) [process, below of=more, yshift=-1.2cm,text width=3cm] {Concatenate all embeddings}; \node (genNoncore) [process, right of=more, xshift=3.5cm,text width=4cm, align=center] {factorize.py:\\ block\_factorize() \\ Solve noncore blocks}; \node (noncore) [io, below of=genNoncore,  yshift=-0.2cm, minimum width=2cm, inner xsep=-14pt,  align=center] {Noncore  Embeddings}; \node (vec) [io, below of=concat, yshift=0.2cm] {Save to .vec}; \node (eval) [process, xshift=3cm, right of=vec, align=center] {evaluate.py \\ Evaluate}; \node (testset) [io, xshift=3cm, right of=eval, xshift=-1cm, align=center] {7 datasets}; \draw [arrow] (corpus) --  (gramcount); \draw [arrow] (gramcount) --  (bigram); \draw [arrow] (bigram) --  (factorize); \draw [arrow] (factorize) --  (core); \draw [arrow] (factorize) --  (more); \draw [arrow, transform canvas={xshift=-0.7em}] ([xshift=2.8em]core) -- (genNoncore); \draw [arrow] (more) -- node[anchor=east] {yes} (concat); \draw [arrow] (more) -- node[anchor=south] {no} (genNoncore); \draw [arrow] (concat) -- (vec); \draw [arrow] (genNoncore) -- (noncore); \draw [arrow] (noncore) |- ([yshift=0.5em]concat); \draw [arrow] (core.east) |- ([yshift=-1.5em]concat); \draw [arrow] (vec) -- (eval); \draw [arrow] (testset) -- (eval);
\tikzstyle{process} = [rectangle, minimum width=3cm, minimum height=1cm, text centered, text width=3cm, draw=black, fill=orange!30] \end{tikzpicture} \vspace{-1ex}
\caption{Toolbox Architecture}
\vspace{-1.5ex}
\end{figure}
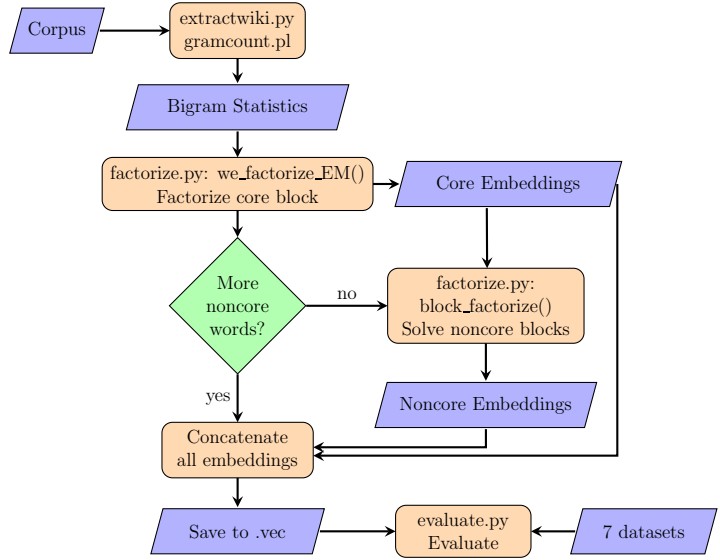

\begin{enumerate}[topsep=2pt,itemsep=-1ex]
\item \texttt{extractwiki.py }first receives a Wikipedia snapshot as input;
it then removes non-textual elements, non-English words and punctuation;
after converting all letters to lowercase, it finally produces a clean
stream of English words;
\item \texttt{gramcount.pl} counts the frequencies of either unigrams or
bigrams in a word stream, and saves them into a file. In the unigram
mode (\texttt{-m1}), unigrams that appear less than certain frequency
threshold are discarded. In the bigram mode (\texttt{-m2}), each pair
of words in a text window (whose size is specified by \texttt{-n})
forms a bigram. Bigrams starting with the same leading word are grouped
together in a row, corresponding to a row in matrices $\boldsymbol{H}$
and $\boldsymbol{G}$;
\item \texttt{factorize.py }is the core module that learns embeddings from
a bigram frequency file generated by \texttt{gramcount.pl}. A user
chooses to split the vocabulary into a set of core words and a few
sets of noncore words. \texttt{factorize.py} can: 1) in function \texttt{we\_factorize\_EM()},
do BCD on the PMI submatrix of core-core bigrams, yielding \emph{core
embeddings}; 2) given the core embeddings obtained in 1), in \texttt{block\_factorize()},
do a weighted ridge regression w.r.t. \emph{noncore embeddings} to
fit the PMI submatrices of core-noncore bigrams. The Tikhonov regularization
coefficient $\mu_{i}$ for a whole noncore block can be specified
by \texttt{-t}. A good rule-of-thumb for setting $\mu_{i}$ is to
increase $\mu_{i}$ as the word frequencies decrease, i.e., give more
penalty to rarer words, since the corpus contains insufficient information
of them;
\item \texttt{evaluate.py }evaluates a given set of embeddings on 7 commonly
used testsets, including 5 similarity tasks and 2 analogy tasks.
\end{enumerate}

\section{Implementation and Empirical Results}

\subsection{Implementation Details}

The Python scripts use Numpy for the matrix computation. Numpy automatically
parallelizes the computation to fully utilize a multi-core machine.

The Perl script \texttt{gramcount.pl} implements an embedded C++ engine
to speed up the processing with a smaller memory footprint.

\subsection{Empirical results}

Our competitors include: \textbf{word2vec}, \textbf{PPMI} and \textbf{SVD
}in Hyperwords, \textbf{GloVe}, \textbf{Singular} and \textbf{Sparse}.
In addition, to show the effect of Tikhonov regularization on ``PSDVec'',
evaluations were done on an unregularized PSDVec (by passing ``\texttt{-t
0}'' to \texttt{factorize.py}), denoted as \textbf{PSD-unreg}. All
methods were trained on an 12-core Xeon 3.6GHz PC with 48 GB of RAM. 

We evaluated all methods on two types of testsets. The first type
of testsets are shipped with our toolkit, consisting of 7 word similarity
tasks and 2 word analogy tasks (Luong's Rare Words is excluded due
to many rare words contained). 7 out of the 9 testsets are used in
\cite{tacl}. The hyperparameter settings of other methods and evaluation
criteria are detailed in \cite{tacl,cca2,sparsecoding1}. The other
2 tasks are TOEFL Synonym Questions (\textbf{TFL}) \cite{toefl} and
Rubenstein \& Goodenough (\textbf{RG}) dataset \cite{rg65}. For these
tasks, all 7 methods were trained on the Apri 2015 English Wikipedia.
All embeddings except ``Sparse'' were 500 dimensional. ``Sparse''
needs more dimensionality to cater for vector sparsity, so its dimensionality
was set to 2500. It used the embeddings of word2vec as the input.
In analogy tasks \texttt{Google} and \texttt{MSR}, embeddings were
evaluated using 3CosMul \cite{3cos}{\scriptsize{}.} The embedding
set of PSDVec for these tasks contained 180,000 words, which was trained
using the blockwise online learning procedure described in Section
5, based on 25,000 core words.

The second type of testsets are 2 practical NLP tasks for evaluating
word embedding methods as used in \cite{turian}, i.e., Named Entity
Recognition (\textbf{NER}) and Noun Phrase Chunking (\textbf{Chunk}).
Following settings in \cite{turian}, the embeddings for NLP tasks
were trained on Reuters Corpus, Volume 1 \cite{rcv1}, and the embedding
dimensionality was set to 50 (``Sparse'' had a dimensionality of
500). The embedding set of PSDVec for these tasks contained 46,409
words, based on 15,000 core words.\addtolength{\tabcolsep}{-2pt}

\begin{table*}[tp]
\begin{centering}
\caption{Performance of each method across 9 tasks.}
\vspace{5pt}
\par\end{centering}

\centering{}{\scriptsize{}}%
\begin{minipage}[t]{1\columnwidth}%
\begin{center}
{\scriptsize{}}%
\begin{tabular}{|c|c|c|c|c|c|c|c|c|c|c|c|c|}
\hline 
 & \multicolumn{7}{c|}{{\scriptsize{}Similarity Tasks}} & \multicolumn{2}{c|}{{\scriptsize{}Analogy Tasks}} & \multicolumn{2}{c|}{{\scriptsize{}NLP Tasks}} & {\scriptsize{}Avg.}\tabularnewline
\hline 
{\scriptsize{}Method } & {\scriptsize{}WS } & {\scriptsize{}WR } & {\scriptsize{}MEN } & {\scriptsize{}Turk } & {\scriptsize{}SL } & {\scriptsize{}TFL} & {\scriptsize{}RG} & {\scriptsize{}Google } & {\scriptsize{}MSR} & {\scriptsize{}NER} & {\scriptsize{}Chunk} & \tabularnewline
\hline 
\hline 
{\scriptsize{}word2vec } & {\scriptsize{}74.1 } & {\scriptsize{}54.8 } & {\scriptsize{}73.2 } & \textbf{\scriptsize{}68.0 } & {\scriptsize{}37.4 } & {\scriptsize{}85.0} & {\scriptsize{}81.1} & \textbf{\scriptsize{}72.3} & \textbf{\scriptsize{}63.0} & \textbf{\scriptsize{}84.8} & {\scriptsize{}94.8} & {\scriptsize{}71.7}\tabularnewline
\hline 
{\scriptsize{}PPMI } & {\scriptsize{}73.5 } & {\scriptsize{}67.8 } & {\scriptsize{}71.7 } & {\scriptsize{}65.9 } & {\scriptsize{}30.8 } & {\scriptsize{}70.0} & {\scriptsize{}70.8} & {\scriptsize{}52.4} & {\scriptsize{}21.7} & {\scriptsize{}N.A.\raisebox{0.5ex}{*}} & {\scriptsize{}N.A.}\footnote{{\scriptsize{}These two experiments are impractical for ``PPMI'',
as they use embeddings as features, and the dimensionality of a PPMI
embedding equals the size of the vocabulary, which is over 40,000. }} & {\scriptsize{}58.3}\tabularnewline
\hline 
{\scriptsize{}SVD } & {\scriptsize{}69.2 } & {\scriptsize{}60.2 } & {\scriptsize{}70.7 } & {\scriptsize{}49.1 } & {\scriptsize{}28.1 } & {\scriptsize{}57.5} & {\scriptsize{}71.8} & {\scriptsize{}24.0} & {\scriptsize{}11.3} & {\scriptsize{}81.2} & {\scriptsize{}94.1} & {\scriptsize{}56.1}\tabularnewline
\hline 
{\scriptsize{}GloVe } & {\scriptsize{}75.9 } & {\scriptsize{}63.0 } & {\scriptsize{}75.6 } & {\scriptsize{}64.1 } & {\scriptsize{}36.2 } & {\scriptsize{}87.5} & {\scriptsize{}77.0} & {\scriptsize{}54.4} & {\scriptsize{}43.5} & {\scriptsize{}84.5} & {\scriptsize{}94.6} & {\scriptsize{}68.8}\tabularnewline
\hline 
{\scriptsize{}Singular } & {\scriptsize{}76.3 } & \textbf{\scriptsize{}68.4}{\scriptsize{} } & {\scriptsize{}74.7 } & {\scriptsize{}58.1 } & {\scriptsize{}34.5 } & {\scriptsize{}78.8} & {\scriptsize{}80.7} & {\scriptsize{}50.8} & {\scriptsize{}39.9} & {\scriptsize{}83.8} & {\scriptsize{}94.8} & {\scriptsize{}67.3}\tabularnewline
\hline 
{\scriptsize{}Sparse } & {\scriptsize{}74.8} & {\scriptsize{}56.5} & {\scriptsize{}74.2} & {\scriptsize{}67.6 } & {\scriptsize{}38.4} & \textbf{\scriptsize{}88.8} & {\scriptsize{}81.6} & {\scriptsize{}71.6} & {\scriptsize{}61.9} & {\scriptsize{}78.8} & \textbf{\scriptsize{}94.9} & {\scriptsize{}71.7}\tabularnewline
\hline 
{\scriptsize{}PSDVec } & \textbf{\scriptsize{}79.2}{\scriptsize{} } & {\scriptsize{}67.9 } & \textbf{\scriptsize{}76.4}{\scriptsize{} } & {\scriptsize{}67.6 } & \textbf{\scriptsize{}39.8}{\scriptsize{} } & {\scriptsize{}87.5} & \textbf{\scriptsize{}83.5} & {\scriptsize{}62.3} & {\scriptsize{}50.7} & {\scriptsize{}84.7} & {\scriptsize{}94.7} & \textbf{\scriptsize{}72.2}\tabularnewline
\hline 
{\scriptsize{}PSD-unreg}\footnote{{\scriptsize{}``PSDVec'' with all Tikhonov regularization coefficients
$\mu_{i}=0$, i.e., unregularized.}} & {\scriptsize{}78.6} & {\scriptsize{}66.3} & {\scriptsize{}75.3} & {\scriptsize{}67.5} & {\scriptsize{}37.2} & {\scriptsize{}85.0} & {\scriptsize{}79.9} & {\scriptsize{}59.8} & {\scriptsize{}46.8} & {\scriptsize{}84.7} & {\scriptsize{}94.5} & {\scriptsize{}70.5}\tabularnewline
\hline 
\end{tabular}
\par\end{center}%
\end{minipage}{\scriptsize{}\vspace{-1ex}
}{\scriptsize \par}
\end{table*}
\addtolength{\tabcolsep}{2pt}Table 1 above reports the performance
of 7 methods on 11 tasks. The last column reports the average score.
``PSDVec'' performed stably across the tasks, and achieved the best
average score. On the two analogy tasks, ``word2vec'' performed
far better than all other methods (except ``Sparse'', as it was
derived from ``word2vec''), the reason for which is still unclear.
On NLP tasks, most methods achieved close performance. ``PSDVec''
outperformed ``PSD-unreg'' on all tasks.

To compare the efficiency of each method, we presented the training
time of different methods across 2 training corpora in Table 2. Please
note that the ratio of running time is determined by a few factors
together: the ratio of vocabulary sizes ($180000/46409\approx4$),
the ratio of vector lengths ($500/50=10$), the language efficiency,
and the algorithm efficiency. We were most interested in the algorithm
efficiency. To reduce the effect of different language efficiency
of different methods, we took the ratio of the two training time to
measure the scalability of each algorithm.

From Table 2, we can see that ``PSDVec'' exhibited a competitive
absolute speed, considering the inefficiency of Python relative to
C/C++. The scalability of ``PSDVec'' ranked the second best, worse
than ``Singular'' and better than ``word2vec''. 

The reason that ``PPMI'' and ``SVD'' (based on ``PPMI'') were
so slow is that ``hyperwords'' employs an external sorting command,
which is extremely slow on large files. The reason for the poor scalability
of ``Sparse'' is unknown.

\begin{table*}[tp]
\begin{centering}
\caption{Training time (minutes) of each method across 2 training corpora.}
\vspace{5pt}
\par\end{centering}

\centering{}{\scriptsize{}}%
\begin{minipage}[t]{1\columnwidth}%
\begin{center}
{\scriptsize{}}%
\begin{tabular}{|c|c|c|c|c|}
\hline 
{\scriptsize{}Method } & {\scriptsize{}Language} & {\scriptsize{}Wikipedia} & {\scriptsize{}RCV1 } & {\scriptsize{}Ratio}\tabularnewline
\hline 
\hline 
{\scriptsize{}word2vec } & {\scriptsize{}C} & {\scriptsize{}249} & {\scriptsize{}15} & {\scriptsize{}17}\tabularnewline
\hline 
{\scriptsize{}PPMI } & {\scriptsize{}Python} & {\scriptsize{}2196} & {\scriptsize{}57} & {\scriptsize{}39}\tabularnewline
\hline 
{\scriptsize{}SVD } & {\scriptsize{}Python} & {\scriptsize{}2282} & {\scriptsize{}58} & {\scriptsize{}39}\tabularnewline
\hline 
{\scriptsize{}GloVe } & {\scriptsize{}C} & {\scriptsize{}229} & {\scriptsize{}6} & {\scriptsize{}38}\tabularnewline
\hline 
{\scriptsize{}Singular } & {\scriptsize{}C++} & \textbf{\scriptsize{}183} & {\scriptsize{}26} & {\scriptsize{}7}\tabularnewline
\hline 
{\scriptsize{}Sparse } & {\scriptsize{}C++} & {\scriptsize{}1548} & \textbf{\scriptsize{}1} & {\scriptsize{}1548}\tabularnewline
\hline 
\textbf{\scriptsize{}PSDVec } & {\scriptsize{}Python} & {\scriptsize{}463} & {\scriptsize{}34} & {\scriptsize{}14}\tabularnewline
\hline 
\emph{\scriptsize{}PSD-core}\footnote{This is the time of generating the core embeddings only, and is not
compariable to other methods.} & \emph{\scriptsize{}Python} & \emph{\scriptsize{}137} & \emph{\scriptsize{}31} & \emph{\scriptsize{}4}\tabularnewline
\hline 
\end{tabular}
\par\end{center}%
\end{minipage}{\scriptsize{}\vspace{-1ex}
}{\scriptsize \par}
\end{table*}

Table 3 shows the time and space efficiency of the incremental learning
(``PSD-noncore'' for noncore words) and MF-based learning (``PSD-core''
for core words) on two corpora. The memory is halved using incremental
learning, and is constant as the vocabulary size increases. Remind
that the asymptotic per-word time complexity of ``PSD-noncore''
is $cd^{2}/\mu n^{2}$ of that of ``PSD-core'', in which typically
$\mu>20$. As embedding dimensionality $d$ on Wikipedia is 10 times
of that on RCV1, the speedup rate on the Wikipedia corpus is only
around $1/12$ of that on the RCV1 corpus\footnote{According to the expression $cd^{2}/\mu n^{2}$, the speedup rate
on Wikipedia should be $1/60$ of that on RCV1. But some common overhead
of Numpy matrix operations is more prominent on the smaller matrices
when $d$ is small, which reduces the speedup rate on smaller $d$.
Hence the ratio of the two speedup rates is $1/12$ in practice.}.

\addtolength{\tabcolsep}{-4pt}
\begin{table*}
\begin{centering}
\caption{Efficiency of incremental learning of PSDVec.}
\vspace{5pt}
\par\end{centering}

\centering{}{\scriptsize{}}%
\begin{tabular}{|c|c|c|c|c|c|c|c|c|c|c|}
\hline 
 & \multicolumn{5}{c|}{{\scriptsize{}Wikipedia ($c=25000,d=500$)}} & \multicolumn{5}{c|}{{\scriptsize{}RCV1 ($c=15000,d=50$)}}\tabularnewline
\hline 
{\scriptsize{}Method} & {\scriptsize{}words} & {\scriptsize{}time (m)} & {\scriptsize{}RAM (G)} & {\scriptsize{}words/m} & {\scriptsize{}speedup} & {\scriptsize{}words} & {\scriptsize{}time (m)} & {\scriptsize{}RAM (G)} & {\scriptsize{}words/m} & {\scriptsize{}speedup}\tabularnewline
\hline 
\hline 
{\scriptsize{}PSD-core} & {\scriptsize{}25000} & {\scriptsize{}137} & {\scriptsize{}44} & {\scriptsize{}182} & {\scriptsize{}1} & {\scriptsize{}15000} & {\scriptsize{}31} & {\scriptsize{}15} & {\scriptsize{}500} & {\scriptsize{}1}\tabularnewline
\hline 
{\scriptsize{}PSD-noncore} & {\scriptsize{}155000} & {\scriptsize{}326} & {\scriptsize{}22} & {\scriptsize{}375} & {\scriptsize{}2.1} & {\scriptsize{}31409} & {\scriptsize{}2.5} & {\scriptsize{}8} & {\scriptsize{}12500} & {\scriptsize{}25}\tabularnewline
\hline 
\end{tabular}{\scriptsize{}\vspace{-1ex}
}{\scriptsize \par}
\end{table*}

\addtolength{\tabcolsep}{4pt}%

\section{Illustrative Example: Training on English Wikipedia}

In this example, we train embeddings on the English Wikipedia snapshot
in April 2015. The training procedure goes as follows:

\begingroup\footnotesize
\begin{enumerate}[topsep=2pt,itemsep=-0.3ex]
\item Use \texttt{extractwiki.py} to cleanse a Wikipedia snapshot, and
generate \texttt{cleanwiki.txt}, which is a stream of 2.1 billion
words;
\item Use \texttt{gramcount.pl} with \texttt{cleanwiki.txt} as input, to
generate \texttt{top1grams-wiki.txt};
\item Use \texttt{gramcount.pl} with \texttt{top1grams-wiki.txt} and \texttt{cleanwiki.txt}
as input, to generate \texttt{top2grams-wiki.txt};
\item Use \texttt{factorize.py} with \texttt{top2grams-wiki.txt }as input,
to obtain 25000 core embeddings, saved into \texttt{25000-500-EM.vec};
\item Use \texttt{factorize.py} with \texttt{top2grams-wiki.txt }and \texttt{25000-500-EM.vec}
as input, and Tikhonov regularization coefficient set to 2, to obtain
55000 noncore embeddings. The word vectors of totally 80000 words
is saved into \texttt{25000-80000-500-BLKEM.vec};
\item Repeat Step 5 twice with Tikhonov regularization coefficient set to
4 and 8, respectively, to obtain extra $50000\times2$ noncore embeddings.
The word vectors are saved into \texttt{25000-130000-500-BLKEM.vec
}and \texttt{25000-180000-500-BLKEM.vec}, respectively;
\item Use \texttt{evaluate.py} to test \texttt{25000-180000-500-BLKEM.vec}.
\end{enumerate}
\endgroup

\section{Conclusions}

We have developed a Python/Perl toolkit \texttt{PSDVec} for learning
word embeddings from a corpus. This open-source cross-platform software
is easy to use, easy to extend, scales up to large vocabularies, and
can learn new words incrementally without re-training the whole vocabulary.
The produced embeddings performed stably on various test tasks, and
achieved the best average score among 7 state-of-the-art methods.

\section*{Acknowledgements}

This research is supported by the National Research Foundation Singapore
under its Interactive Digital Media (IDM) Strategic Research Programme.

\bibliographystyle{plain}

\begin{thebibliography}{10}

\bibitem{lda}
David~M Blei, Andrew~Y Ng, and Michael~I Jordan.
\newblock Latent dirichlet allocation.
\newblock {\em the Journal of machine Learning research}, 3:993--1022, 2003.

\bibitem{sparsecoding1}
Manaal Faruqui, Yulia Tsvetkov, Dani Yogatama, Chris Dyer, and Noah~A. Smith.
\newblock Sparse overcomplete word vector representations.
\newblock In {\em Proceedings of ACL}, 2015.

\bibitem{toefl}
Thomas~K Landauer and Susan~T Dumais.
\newblock A solution to plato's problem: The latent semantic analysis theory of
  acquisition, induction, and representation of knowledge.
\newblock {\em Psychological review}, 104(2):211, 1997.

\bibitem{ppmi}
Omer Levy and Yoav Goldberg.
\newblock Neural word embeddings as implicit matrix factorization.
\newblock In {\em Proceedings of NIPS 2014}, 2014.

\bibitem{tacl}
Omer Levy, Yoav Goldberg, and Ido Dagan.
\newblock Improving distributional similarity with lessons learned from word
  embeddings.
\newblock {\em Transactions of the Association for Computational Linguistics},
  3:211--225, 2015.

\bibitem{3cos}
Omer Levy, Yoav Goldberg, and Israel Ramat-Gan.
\newblock Linguistic regularities in sparse and explicit word representations.
\newblock In {\em Proceedings of CoNLL-2014}, page 171, 2014.

\bibitem{rcv1}
David~D Lewis, Yiming Yang, Tony~G Rose, and Fan Li.
\newblock Rcv1: A new benchmark collection for text categorization research.
\newblock {\em The Journal of Machine Learning Research}, 5:361--397, 2004.

\bibitem{topicvec}
Shaohua Li, Tat-Seng Chua, Jun Zhu, and Chunyan Miao.
\newblock Topic embedding: a continuous representation of documents.
\newblock In {\em Proceedings of the The 54th Annual Meeting of the Association
  for Computational Linguistics (ACL)}, 2016.

\bibitem{psdvec}
Shaohua Li, Jun Zhu, and Chunyan Miao.
\newblock A generative word embedding model and its low rank positive
  semidefinite solution.
\newblock In {\em Proceedings of the 2015 Conference on Empirical Methods in
  Natural Language Processing}, pages 1599--1609, Lisbon, Portugal, September
  2015. Association for Computational Linguistics.

\bibitem{word2vec}
Tomas Mikolov, Ilya Sutskever, Kai Chen, Greg~S Corrado, and Jeff Dean.
\newblock Distributed representations of words and phrases and their
  compositionality.
\newblock In {\em Proceedings of NIPS 2013}, pages 3111--3119, 2013.

\bibitem{glove}
Jeffrey Pennington, Richard Socher, and Christopher~D Manning.
\newblock Glove: Global vectors for word representation.
\newblock {\em Proceedings of the Empirical Methods in Natural Language
  Processing (EMNLP 2014)}, 12, 2014.

\bibitem{rg65}
Herbert Rubenstein and John~B. Goodenough.
\newblock Contextual correlates of synonymy.
\newblock {\em Commun. ACM}, 8(10):627--633, October 1965.

\bibitem{weightedlow}
Nathan Srebro, Tommi Jaakkola, et~al.
\newblock Weighted low-rank approximations.
\newblock In {\em Proceedings of ICML 2003}, volume~3, pages 720--727, 2003.

\bibitem{cca2}
Karl Stratos, Michael Collins, and Daniel Hsu.
\newblock Model-based word embeddings from decompositions of count matrices.
\newblock In {\em Proceedings of ACL}, 2015.

\bibitem{turian}
Joseph Turian, Lev Ratinov, and Yoshua Bengio.
\newblock Word representations: a simple and general method for semi-supervised
  learning.
\newblock In {\em Proceedings of the 48th annual meeting of the association for
  computational linguistics}, pages 384--394. Association for Computational
  Linguistics, 2010.

\end{thebibliography}

\clearpage{}

\section*{Required Metadata}

\section*{Current code version}

Ancillary data table required for subversion of the codebase. Kindly
replace examples in right column with the correct information about
your current code, and leave the left column as it is.

\begin{table}[!h]
\begin{tabular}{|l|>{\centering}m{6cm}|>{\centering}m{7cm}|}
\hline 
\textbf{Nr.}  & \textbf{Code metadata description}  & \tabularnewline
\hline 
C1  & Current code version  & 0.4\tabularnewline
\hline 
C2  & Permanent link to code/repository used of this code version  & https://github.com/askerlee/topicvec\tabularnewline
\hline 
C3  & Legal Code License  & GPL-3.0\tabularnewline
\hline 
C4  & Code versioning system used  & git\tabularnewline
\hline 
C5  & Software code languages, tools, and services used  & Python, Perl, (inline) C++\tabularnewline
\hline 
C6  & Compilation requirements, operating environments \& dependencies  & Python: numpy, scipy, psutils; Perl: Inline::CPP; C++ compiler\tabularnewline
\hline 
C7  & If available Link to developer documentation/manual  & N/A\tabularnewline
\hline 
C8  & Support email for questions  & shaohua@gmail.com\tabularnewline
\hline 
\end{tabular}\caption{Code metadata (mandatory)}
\end{table}

\end{document}